\documentclass[a4paper,conference]{IEEEtran}

\usepackage{amsmath}
\usepackage{algorithm}
\usepackage[noend]{algpseudocode}

\makeatletter
\def\BState{\State\hskip-\ALG@thistlm}
\makeatother
\usepackage{graphicx}
\usepackage{hhline}
\usepackage{multirow}
\ifCLASSINFOpdf
\else
\fi
\usepackage{subcaption}

\usepackage[export]{adjustbox}

\begin{document}
%
\title{Memory-Efficient Deep Salient Object Segmentation Networks on Gridized Superpixels}

\author{\IEEEauthorblockN{Caglar Aytekin, Xingyang Ni, Francesco Cricri, Lixin Fan, Emre Aksu}
\IEEEauthorblockA{Nokia Technologies, Tampere,
Finland \\
Corresponding Author's Email: caglar.aytekin@nokia.com}}


%


\maketitle

\begin{abstract}
Computer vision algorithms with pixel-wise labeling tasks, such as semantic segmentation and salient object detection, have gone through a significant accuracy increase with the incorporation of deep learning. 
Deep segmentation methods slightly modify and fine-tune pre-trained networks that have hundreds of millions of parameters.
In this work, we question the need to have such memory demanding networks for the specific task of salient object segmentation.
To this end, we propose a way to learn a memory-efficient network from scratch by training it only on salient object detection datasets. 
Our method encodes images to gridized superpixels that preserve both the object boundaries and the connectivity rules of regular pixels. 
This representation allows us to use convolutional neural networks that operate on regular grids.
By using these encoded images, we train a memory-efficient network using only 0.048\% of the number of parameters that other deep salient object detection networks have. 
Our method shows comparable accuracy with the state-of-the-art deep salient object detection methods and provides a faster and a much more memory-efficient alternative to them.
Due to its easy deployment, such a network is preferable for applications in memory limited devices such as mobile phones and IoT devices.

\end{abstract}


%
\IEEEpeerreviewmaketitle

\section{Introduction}
{C}{onvolutional} Neural Networks (CNNs) are learning machines that are extensively used by top performing methods in image classification \cite{Krizhevsky,Simonyan,Szegedy,He}.
By the introduction of Fully Convolutional Neural Networks (FCNNs) \cite{Long}, these structures have also proven to constitute the state of the art in pixel-wise classification tasks such as semantic image segmentation and salient object detection.
A typical FCNN relies on a pre-trained CNN that is used for image classification and fine-tunes the CNN's parameters for segmentation task, often adding or replacing some layers.
These pre-trained CNNs usually contain a very large number of parameters, e.g. 138 million for VGG-16 \cite{Simonyan}.
Such large networks require a lot of memory, which makes them challenging to deploy on limited memory devices such as mobile phones.
There have been some efforts to reduce memory requirement of a CNN via pruning \cite{Hanc} or quantizing \cite{Guptac} the weights of the network, however these approaches are post-processing operations on large networks trained on millions of images. 
For some segmentation tasks such as salient object detection, one might question the need for using such a high capacity network in the first place.
It can be argued that such a network might be an overkill for salient object detection and one can achieve reasonable performance by using a much smaller network.
Moreover, object recognition CNNs have greatly reduced resolutions in their final layer activations due to pooling or strided convolution operations throughout the network.
In order to atone for this resolution loss, segmentation networks either introduces additional connections to make use of the localization power of low-middle layers \cite{Long,Hariharan}, or adds a deconvolutional network on top of the CNN \cite{Badrinarayanan, Noh} with unpooling layers. 
Both approaches results into an even more increase in the number of parameters used in the segmentation network. 

\begin{figure}[!t]
\includegraphics[width=0.5\textwidth,left]{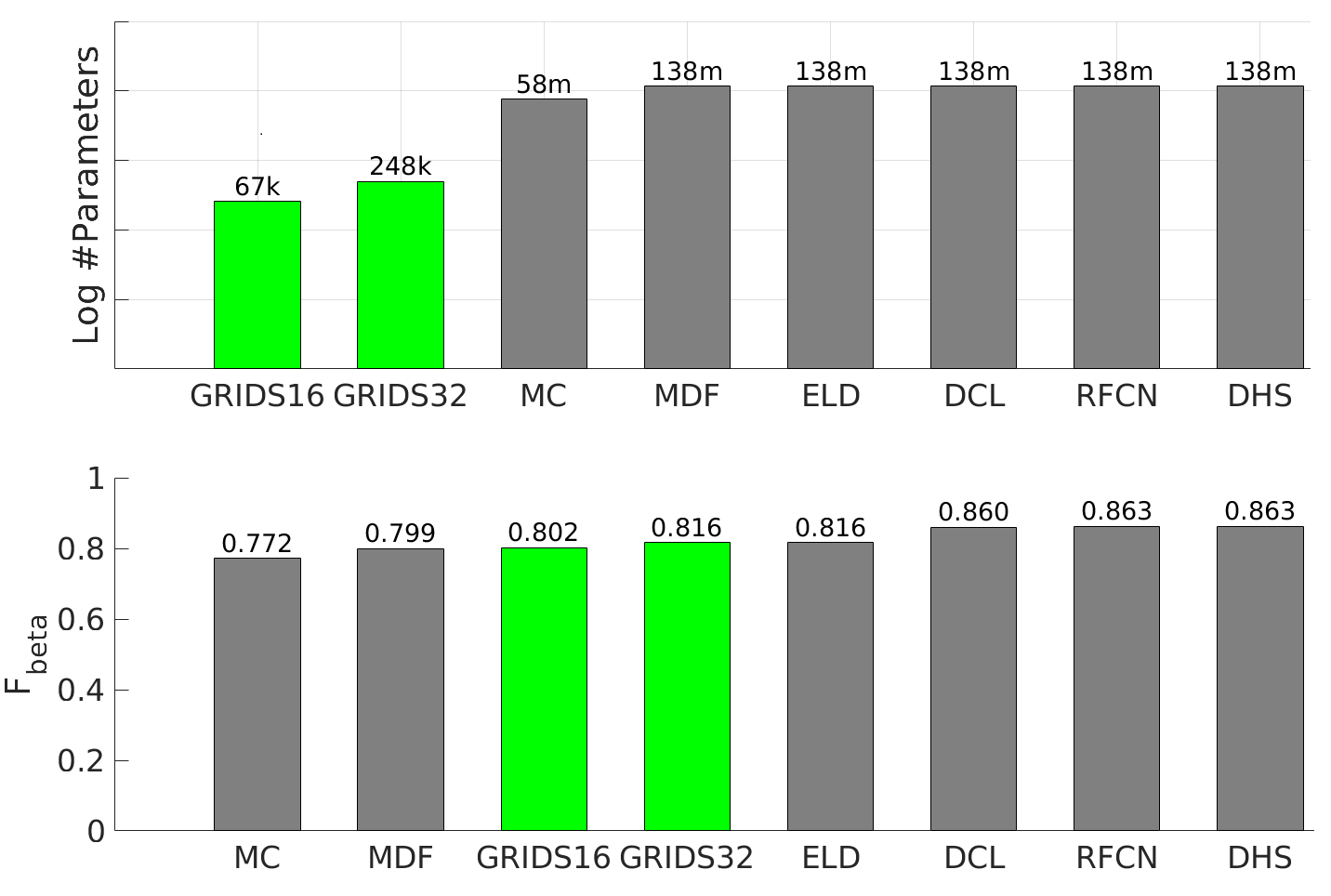}


\caption{Our method (GRIDS) compared to state-of-the-art deep salient object detection methods. (a) Network size comparison (plotted in log-scale), the true number of parameters are written on the corresponding bar. (b) Performance comparison according to $F_\beta$ measure.}
\label{fig0}
\end{figure}

In this paper, we propose a way to overcome two problems of FCNNs mentioned above: requirement of a big pre-trained network and resolution loss because of pooling layers.
To this end, we utilize a memory-efficient deep segmentation network without any pooling layers.
We achieve this by encoding input images via gridized superpixels \cite{Fu}.
This allows us to use low resolution images that accurately encode object edges.
By using these images, we show that it is possible to train a memory-efficient FCNN with a reasonable depth, no pooling layers, yet with large receptive field and comparable performance with state of the art, see Fig. \ref{fig0}.
The contributions of our work are listed as follows:

\begin{itemize}
  \item We propose a way to use FCNNs without any pooling layers or strided convolutions by abstracting input images via gridized superpixels.
  \item The predictions of our network does not suffer from inaccurate object edges.
  \item Our proposed network has less than 67k parameters (about 0.048\% of others).
  \item Our proposed network does not require a pre-trained model and can be trained from scratch by existing pixel-wise classification datasets.
  \item We show that the performance of our method is comparable with state of the art segmentation networks in salient object detection task.
\end{itemize}

The rest of the paper is organized as follows. In Section \ref{RelWork}, the related work is discussed, in Section \ref{PropMet}, the proposed method is described, in \ref{ExpRes}, the experimental results are analyzed. Finally, Section \ref{Conc} concludes the paper and suggests topics for future research.

\section{Related Work} \label{RelWork}
\subsection{Superpixel Gridization}
Superpixel gridization produces over-segmentations that form a regular pixel-like lattice which best preserves the object edges in an image.
There is a small number of studies in this topic which we review shortly next.
The method in \cite{Moore} relies on a boundary cost map which is an inverted edge detection result. Incrementally horizontal and vertical stripes are added to the image where no horizontal and vertical stripes intersect more than once and no two horizontal (or vertical) stripes intersect with each other. In each step the optimal stripe is found by minimizing the boundary cost that the stripe passes through by a min-cut based optimization algorithm.

An extended version of \cite{Moore} was proposed in \cite{Moore2} where the authors use an alternating optimization strategy. The method finds globally optimal solutions to the horizontal and vertical components of the lattice via using a multi-label Markov Random Field, as opposed to the greedy optimization strategy adopted in \cite{Moore}.
In \cite{Li1}, a generic approach is proposed to optimally regularize superpixels extracted by any algorithm. The approach is based on placing dummy nodes between superpixels to satisfy the regularity criterion.

Finally, the method proposed in \cite{Fu}, starts with regular lattice seeds and relocates the seeds -in a search space defined by the initialization- to the locally maximal boundary response. The relocated seeds are considered as superpixel junctions. Next, for each junction pair a path was found that maximize the edge strength on the path. These paths form a superpixel boundary map which results in a regular superpixel grid.   

\subsection{Salient Object Detection}
A salient object is generally defined as the object that visually stands out from the rest of the image, thus is more appealing to the human eye \cite{Borji1}. 

Unsupervised salient object detection methods rely mostly on following saliency assumptions: 1) A salient object has high local or global contrast \cite{Cheng1,Aytekin1}, 2) The boundary of an image is less likely to contain a salient object \cite{Li2,Aytekin2}, 3) The salient object is more likely to be large \cite{Aytekin2}, 4) Regions of similar feature maps have similar saliency \cite{Aytekin3}. 

Prior to deep learning, supervised approaches to salient object detection focused on following tracks: 1) Learning a dense labeling of each region as salient or not \cite{Jiang, Li3}, 2) Learning to rank salient object proposals \cite{Aytekin4}, 3) Learning region affinities for \cite{Aytekin5} end to end salient object segmentation.

Deep learning based approaches to salient object detection either train a network to learn to classify each region in an image separately \cite{Zhao, Li4,Gayoung}, or employ FCNNs to learn a dense pixel-wise labeling for salient object detection \cite{Wang,Li5}. FCNN based models utilize pre-trained networks on other tasks, and employ special tricks to preserve the accurate edges in segmentation results. Next, we propose an FCNN that does not need a pre-trained network, automatically preserves accurate object edges with no pooling layers or strided convolutions, has much fewer parameters than other methods, and is comparable in performance to the state-of-the-art.

\begin{figure}[!t]
\centering
\includegraphics[width=2.5in]{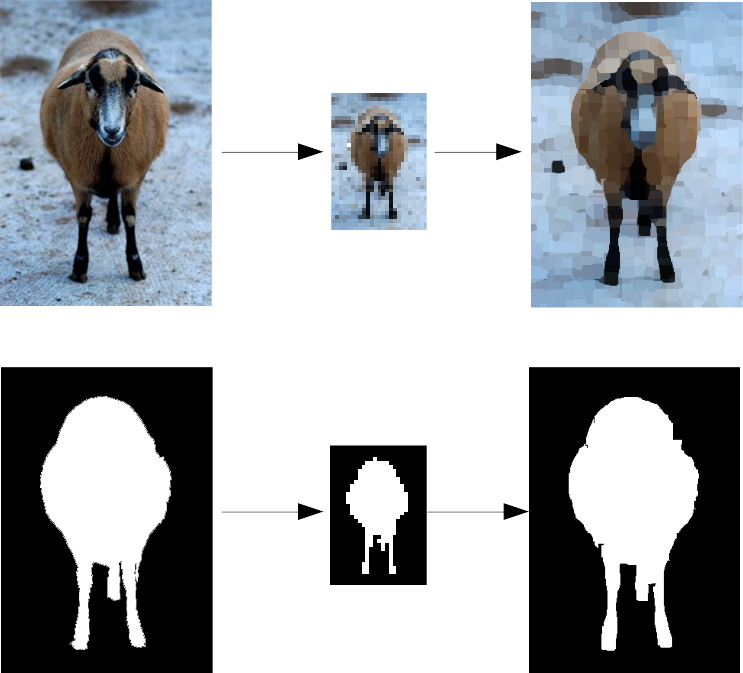}
\caption{From left to right: Original, encoded with \cite{Fu} (950 superpixels) and reconstructed images (first row) and corresponding ground truths (second row).}
\label{fig1}
\end{figure}

\section{Proposed Method} \label{PropMet}
\subsection{Data Preparation}
We use the superpixel extraction method in \cite{Fu} to abstract an input image $ I $ with a small number of homogeneous image regions (superpixels). 
Thanks to the special property of the method in \cite{Fu}, the extracted superpixels form a grid. 
We encode each superpixel with its mean color, thus we end up with a new image $ X $ with low dimensions as follows:
\begin{equation}
X^{(c)}_{i}=\frac{1}{|s_{i}|}\sum_{\{j\}\in s_{i}}I^{(c)}_{j}.
\label{eqn1}
\end{equation}
In Eq. \ref{eqn1}, $ c $ indicates a channel of an image, $ i $ and $ j $ are indices of $ X $ and $ I $ images respectively, $ s_{i} $ is a set of pixels of $ I $ covered in superpixel $ i $, and $ |.| $ is cardinality operation.
We will use this image $ X $ as input to an FCNN. 
For training purposes, we also form a low resolution version of the ground truth label image $ G $ via encoding the mean of the 0 (not salient) and 1 (salient) values in the regions indicated by superpixels similar to the process in Eq. \ref{eqn1}.
In order to have binary values in the low dimensional ground truth $ Y $, we simply threshold the above image by 0.5. Note that it is equivalent to selecting the most common value within a superpixel.
It should be noted that, one can reconstruct an approximation of $ I $ and $ G $ from $ X $ and $ Y $ respectively as follows:
\begin{equation}
\tilde{I}^{(c)}_{j}=X^{(c)}_{i}, j\in s_{i}.
\label{eqn2}
\end{equation}
In Fig. \ref{fig1}, we illustrate original ($ I $), encoded ($ X $) and reconstructed ($ \tilde{I} $) images with corresponding ground truths. 
The encoded image is the input that will be supplied to the FCNN with the encoded ground truth as its label for training. 
As we observe from Fig. \ref{fig1}, even though the superpixel extraction is constrained by the grid structure, it is able to reconstruct the image well by preserving the object edges.

\begin{figure}[!t]
\centering
\includegraphics[width=1.25in]{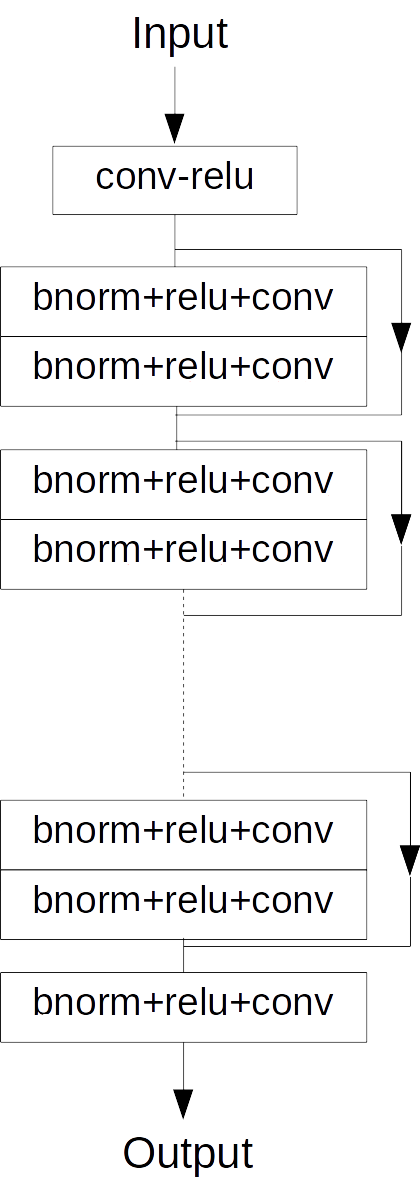}
\caption{Network Architecture.}
\label{fig2}
\end{figure}

\subsection{Network Architecture}
We use a 28-layers deep convolutional network with residual blocks \cite{He}.
In particular, the network has a convolutional (conv) layer with rectified linear unit \cite{Nair} (relu) activation, 13 residual blocks followed by a batch normalization \cite{Ioffe} (bnorm) , relu and conv layer and sigmoid activation. 
Note that we have not applied bnorm layer right before the sigmoid in order to avoid restricting the convolutional output to a small interval.
Each residual block consists of a bnorm-relu-conv-bnorm-relu-conv structure.
The input of a residual block is short-connected to its output.
We use same number of filters in each layer.
The entire network is illustrated in Fig. \ref{fig2}.
The network's inputs and corresponding ground truths are obtained by the procedure described in the previous subsection.
Note that the convolutions are utilized with zero padding and stride 1, so that the input shape is preserved for each convolutional layer.
Moreover, there are no pooling layers in the network architecture in order to avoid any resolution loss.
This is possible because the input resolution is already low and we can use a constant number of convolutional filters throughout the network, whereas prior art networks need to reduce the resolution in order to increase the number of filters.
The receptive field of our network is around 30x30 which is enough to cover the entire input size for an abstraction of an image with 900 superpixels if the abstraction forms a square grid.
Typically the aspect ratio of the superpixel representation varies with the image aspect ratio, however we find the above receptive field enough to accurately detect the salient objects.

\begin{algorithm}
\caption{Test-time implementation}\label{testalgorithm}
\hspace*{\algorithmicindent} \textbf{Input: image $ I $} \\
\hspace*{\algorithmicindent} \textbf{Output: salient segment $ \tilde{S} $ } 
\begin{algorithmic}[1] \\
\text{Encode $ I $ to $ X $ by Eq. \ref{eqn1}} \\
\text{Apply min-max normalization on $ X $} \\
\text{Predict $ S $ from $ X $ via neural network} \\
\text{Reconstruct $ \tilde{S} $ from $ S $ by Eq. \ref{eqn2}}
\end{algorithmic}
\end{algorithm}

\subsection{Training and Testing}
The parameters of the network are optimized in order to minimize the binary cross entropy loss between the output of the network and the ground truth, by treating the sigmoid outputs as probabilities that the corresponding input pixels are salient.
Separate datasets are used for training and validation sets and the model with the best validation error is selected.
For testing, we use entirely different datasets and run the model learned by the training as described above.
During testing, an image is encoded to the low dimensional superpixel grid representation and fed to the network.
It should be noted here that we apply min-max normalization to each input, i.e. we linearly scale the values between 0 and 1.
The output of the network lies in the same grid structure, thus should be converted back, i.e. reconstructed to the original image size.
The reconstruction is simply utilized via replicating the value in each grid node in the image region that the node corresponds to as formulated in Eq. \ref{eqn2}.
The test-time algorithm is given in Algorithm \ref{testalgorithm}.

\section{Experimental Results} \label{ExpRes}

\subsection{Datasets and Evaluation Metrics}
We conducted evaluations on widely used salient object detection datasets. 
MSRA10K \cite{Liu} includes 10000 images that exhibit a simple case with one salient object and clear background, HKU-IS \cite{Li4} includes 4447 images with slightly challenging cases, ECSSD \cite{Yan} includes 1000 relatively complex images , PASCALS \cite{Li6} contains 850 images adopted from PASCAL VOC segmentation dataset, and SOD \cite{Mohavedi} contains 300 images from BSD300 \cite{Martin} segmentation dataset.
We use two most widely used evaluation metrics, mean absolute error (MAE) and F-measure.
For a saliency map $S$ and a binary ground truth $GT$, MAE is defined as follows:
\begin{equation}
MAE=\frac{1}{|S|}\sum_{i=1}^{|S|}|S_i-GT_i|.
\label{eqn3}
\end{equation}

Precision-recall curves are extracted via thresholding the saliency map $S$ at several values $\tau$ and plotting the precision and recall values which are calculated as follows:

\begin{equation}
PRE_\tau=\frac{|S>\tau|\cap|GT|}{|S>\tau|},  REC_\tau=\frac{|S>\tau|\cap|GT|}{|GT|}.
\label{eqn4}
\end{equation}

The F-measure is used to obtain a global evaluation of the precision recall curve and is obtained as follows.

\begin{equation}
F_\beta=\frac{(1+\beta^2)PRE_\tau \times REC_\tau}{\beta^2PRE_\tau+REC_\tau}.
\label{eqn5}
\end{equation}

It is widely adopted in salient object detection literature to chose $\beta^2$ to be $0.3$ and use an adaptive thresholding where $\tau$ equals twice the mean saliency in saliency map. \cite{Borji1}.

\subsection{Implementation}
Our network is based on the publicly available Keras with Theano backend.
Network parameters are initialized by Xavier's method \cite{Xavier}. 
We use Nesterov Adam optimizer with an initial learning rate of $\mu=0.002$.
We utilize a number of superpixel granularities for augmenting the input data. 
In particular, we use $\{900,925,950,975,1000\}$ number of superpixels.
Therefore, we have 5 different encoded image for each original image.
This is only done for training and validation data for data augmentation.
During test stage, we stick to 950 superpixels ,simply because it is the median of the above set, for evaluation.
Unlike other methods, our network is trained from scratch, hence it seems like it needs more data to be trained on, but in reality it is trained on less data if we consider also the pre-training data in prior art works. 
Thus, we use largest datasets DUT-OMRON, HKU-IS and MSRA10K datasets for training and SOD for validation.
Further data augmentation is employed via randomly flipping the input and labels in horizontal direction.
We use a batch size of 20.
The network was run for 5 million iterations and the model that gives the best validation set accuracy was selected.
We evaluate 2 different variants of the network: one with 16 filters at each layer and one with 32.
We call the methods GRIDS16 and GRIDS32 respectively.
The networks that GRIDS16 and GRIDS32 use have 67k and 248k parameters respectively.

\begin{table}
  \caption{Comparison with State-of-the-art}
  \centering
  \renewcommand{\arraystretch}{1.2}
  \begin{tabular}{|p{1cm}|c|c|c|c|c|c|c|}
    \hline
    \multirow{2}{2cm}{\textbf{Method}} & \multicolumn{2}{c|}{\textbf{PASCALS}} & \multicolumn{2}{c|}{\textbf{ECSSD}} & \multicolumn{2}{c|}{\textbf{Avg. Perf.}}   \\
    \cline{2-7}
    & \textbf{MAE} & \textbf{$F_\beta$} & \textbf{MAE} & \textbf{$F_\beta$} & \textbf{MAE} & \textbf{$F_\beta$}  \\
    \hline
    
    CHM & 0.222 & 0.631 & 0.195 & 0.722 & 0.209 & 0.677  \\ \hline    
    RC & 0.225 & 0.640 & 0.187 & 0.741 & 0.206 & 0.691 \\ \hline
    DSR & 0.204 & 0.646 & 0.173 & 0.737 & 0.189 & 0.692  \\ \hline
    EQCUT & 0.217 & 0.670 & 0.174 & 0.765 & 0.196 & 0.718  \\ \hline
    DRFI & 0.221 & 0.679 & 0.166 & 0.787 & 0.194 & 0.733 \\ \hline
    MC & 0.147 & 0.721 & 0.107 & 0.822 & 0.127 & 0.772  \\ \hline
    MDF & 0.145 & 0.764 & 0.108 & 0.833 & 0.127 & 0.799   \\ \hline
    \textbf{GRIDS16} & \textbf{0.171} & \textbf{0.781} & \textbf{0.085} & \textbf{0.823} & \textbf{0.128} & \textbf{0.802} \\ \hline    
        \textbf{GRIDS32} & \textbf{0.166} & \textbf{0.793} & \textbf{0.080} & \textbf{0.839} & \textbf{0.123} & \textbf{0.816} \\ \hline    ELD & 0.121 & 0.767 & 0.098 & 0.865 & 0.110 & 0.816  \\ \hline
    DCL & 0.108 & 0.822 & 0.071 & 0.898 & 0.090 & 0.860 \\ \hline
    RFCN & 0.118 & 0.827 & 0.097 & 0.898 & 0.108 & 0.863  \\ \hline
    DHS & 0.091 & 0.820 & 0.061 & 0.905 & 0.076 & 0.863  \\ \hline
    DSS & 0.080 & 0.830 & 0.052 & 0.915 & 0.066 & 0.873  \\ \hline
  \end{tabular}
  \label{table1}

\end{table}

\subsection{Comparison with State-of-the-art}
We compare our approach with 4 unsupervised methods: RC \cite{Cheng1}, CHM \cite{Li3}, DSR \cite{Li2}, EQCUT \cite{Aytekin2}, and 8 supervised methods: DRFI \cite{Jiang}, MC \cite{Zhao}, ELD \cite{Gayoung}, MDF \cite{Li4}, RFCN \cite{Wang}, DHS \cite{Liu2}, DCL \cite{Li5} and DSS \cite{Hou}. 
In Table \ref{table1}, we share results for ECSSD and PASCALS datasets as these are the only datasets used for testing in all methods. 
The ordering of the methods is made according to ascending $F_\beta$ measure.
As one can observe from Table \ref{table1}, both variants of our method GRIDS can achieve better $F_\beta$ measure and MAE than that of all unsupervised methods and three supervised methods: DRFI, MC and MDF. 
Out of these methods MC and MDF are deep learning based and use around 58 and 138 million parameters respectively. 
Other methods that outperform our method are all deep learning based algorithms and use more than 138 million parameters - VGG-16 models with additional layers/connections. 
Yet, our models GRIDS16 and GRIDS32 only use around 67 and 248 thousand parameters respectively, which corresponds to respectively 0.048\% and 0.18\% of other methods and still achieve a comparable accuracy with the state of the art. 
The number of parameters each deep learning based method use and run times with used GPUs are given in Table \ref{table2}. 
Our method is the one with least memory requirement and fastest run-time. 
At this point, one should note that the superpixel extraction time is not included in the above table. With the method of \cite{Fu}, this takes around an additional 0.5 seconds for an image of size 300x400 for superpixel granularity 950.

\begin{table}[!t]
\renewcommand{\arraystretch}{1.3}

\caption{Complexity of Deep Learning Methods}
\label{table2}
\centering
\begin{tabular}{|c||c||c||c|}
\hline
Method & \#Parameters & Run-time (sec.) & GPU\\
\hline
MC & 58m & 2.38 & Titan Black\\\hline
MDF & 138m & 8 & Titan Black\\\hline
ELD & 138m & 0.5 & Titan Black\\\hline
DCL & 138m & 1.5 & Titan Black\\\hline
RFCN & 138m & 4.6 & Titan X\\\hline
DHS & 138m & 0.04 & Titan Black\\\hline
DSS & 138m & 0.08 & Titan X\\\hline
\textbf{GRIDS16} & \textbf{67k} & \textbf{0.02} & \textbf{GTX 1080}\\\hline
\textbf{GRIDS32} & \textbf{248k} & \textbf{0.03} & \textbf{GTX 1080}\\\hline

\end{tabular}
\end{table}

\subsection{Analysis and Variants}
In this section, we investigate the impact of several factors to our method's performance.
First, we evaluate the performance robustness to different superpixel granularities. 
We report the test performance when employing 900, 950 and 1000 number of superpixels in Table \ref{table3}.
The experiments are made with GRIDS32 model.
It can be observed that the resolution change in this interval has an insignificant impact on our method's performance and does not alter the ranking in Table \ref{table1}.

\begin{table}[!t]
\renewcommand{\arraystretch}{1.3}
\caption{Robustness to Resolution}
\label{table3}
\centering
\begin{tabular}{|c||c||c||c||c|}
\hline
    \multirow{2}{2cm}{\textbf{Superpixel No.}} & \multicolumn{2}{c|}{\textbf{PASCALS}} & \multicolumn{2}{c|}{\textbf{ECSSD}}   \\
    \cline{2-5}
     &\textbf{MAE} & \textbf{$F_\beta$} & \textbf{MAE} & \textbf{$F_\beta$}   \\
    \hline

    900 & 0.169 & 0.787 & 0.080 & 0.838   \\ \hline    
    950 & 0.166 & 0.793 & 0.080 & 0.839 \\ \hline
    1000 & 0.167 & 0.792 & 0.080 & 0.840  \\ \hline

\end{tabular}
\end{table}

Next, we have tried to improve the performance via combination of segmentation results from all resolutions via majority voting. 
Experiments are made with GRIDS32 model.
It can be observed from Table \ref{table3} that multi-resolution approach (GRIDSM) results into a notable performance improvement in both MAE and $F_\beta$ measure.
The rank of GRIDSM is the same with GRIDS for MAE, but it beats one more deep learning method (ELD) in $F_\beta$ measure in Table \ref{table1}.

\begin{table}[!t]
\renewcommand{\arraystretch}{1.3}
\caption{Improvement via Multi-resolution Approach}
\label{table4}
\centering
\begin{tabular}{|c||c||c||c||c|}
\hline
    \multirow{2}{2cm}{\textbf{Method}} & \multicolumn{2}{c|}{\textbf{PASCALS}} & \multicolumn{2}{c|}{\textbf{ECSSD}}   \\
    \cline{2-5}
     &\textbf{MAE} & \textbf{$F_\beta$} & \textbf{MAE} & \textbf{$F_\beta$}   \\
    \hline

    GRIDS (950) & 0.166 & 0.793 & 0.080 & 0.839    \\ \hline    
    GRIDSM & 0.164 & 0.800 & 0.075 & 0.851 \\ \hline

\end{tabular}
\end{table}

One might argue that a natural baseline related to our network is (a) plain downsampling of the datasets, (b) training a network on the downsampled images and ground truths (c) upsampling results to evaluate the performance. This would highlight the performance upgrade of dimension reduction and later reconstruction with gridized superpixel encoding compared to plain downsampling and upsampling. To make this comparison, we train a network with exactly the same structure as described in the text, only this time we train the network with downsampled images and ground truths. Ground truths were again binarized via thresholding with 0.5. Augmentation with scale was similarly performed by defining downsampled dimensions to result into around 900,925,950,975 and 1000 number of pixels while preserving the aspect ratio. During test time we have again utilized only the 950 granularity. Bicubic downsampling and upsampling is used. Experiments are made with GRIDS32 model. The comparison in \ref{table5} clearly indicates the improvement of superpixel gridization encoding scheme over plain downsampling. Especially the improvement in $F_\beta$ measure is dramatic; up to a 13\% relative improvement.

\begin{table}[!t]
\renewcommand{\arraystretch}{1.3}
\caption{Encoding Strategy Comparison}
\label{table5}
\centering
\begin{tabular}{|c||c||c||c||c|}
\hline
    \multirow{2}{2cm}{\textbf{Encoding Type}} & \multicolumn{2}{c|}{\textbf{PASCALS}} & \multicolumn{2}{c|}{\textbf{ECSSD}}   \\
    \cline{2-5}
     &\textbf{MAE} & \textbf{$F_\beta$} & \textbf{MAE} & \textbf{$F_\beta$}   \\
    \hline

    Downsampling & 0.172 & 0.732 & 0.096 & 0.741   \\ \hline    
    GRIDS & 0.166 & 0.793 & 0.080 & 0.839 \\ \hline

\end{tabular}
\end{table}

As we have previously mentioned, since our method is trained from scratch we obviously need more data to be trained on. That is why we use largest datasets DUT-OMRON, HKU-IS and MSRA10K for training.
We would like to emphasize here that the majority of other works use only MSRA10K for training and validation purposes for fine-tuning the pre-trained network they use. 
For transfer learning, fine-tuning with little number of data is known to give satisfactory results. 
Since our network is trained from scratch, such small data is not enough to train a network that gives satisfactory generalization.
Moreover, we do not possess the advantage of starting with a pre-trained network on millions of images for object detection -as others do- which is clearly expected to contribute to salient object detection performance acting as a top-down prior, i.e. using the semantically higher level information of object recognition for detecting the salient object.
Therefore, we argue the fairness of a comparison where we use only MSRA10k for our method.
Yet, in order to give a complete set of experiments, we have also trained our model with other methods' training and validation sets (partitions on MSRA10K dataset) and we report the test results in Table \ref{table6}.
Experiments are made with GRIDS32 model.
Clearly, the model trained on MSRA10K is inferior to the one trained on MSRA10K, HKU-IS and DUT-OMRON.

\begin{table}[!t]
\renewcommand{\arraystretch}{1.3}
\caption{Impact of Training Data}
\label{table6}
\centering
\begin{tabular}{|c||c||c||c||c|}
\hline
    \multirow{2}{2cm}{\textbf{Training Data}} & \multicolumn{2}{c|}{\textbf{PASCALS}} & \multicolumn{2}{c|}{\textbf{ECSSD}}   \\
    \cline{2-5}
     &\textbf{MAE} & \textbf{$F_\beta$} & \textbf{MAE} & \textbf{$F_\beta$}   \\
    \hline

    MSRA10k & 0.190 & 0.752 & 0.096 & 0.814 \\ \hline
    MSRA10k+HKUIS+DUTOMRON & 0.166 & 0.793 & 0.080 & 0.839   \\ \hline    

\end{tabular}
\end{table}

\section{Conclusion and Future Work}  \label{Conc}
We have presented a deep, fast and memory-efficient method for salient object segmentation operating on encoded images with gridized superpixels. With the boundary preserving gridized superpixel encoding, we also do not suffer from blurry object boundaries. Moreover, the network does not employ any pooling layer, thus further resolution loss is prevented. This also eliminates the need of tricks such as additional connections and layers to atone for the resolution loss. We have shown that our method can outperform some deep learning based methods and shows comparable accuracy with others while having only 0.048\% of their parameters. With only 430 KB memory, the network is extremely easy to deploy to any device. This especially makes the method preferable considering applications in mobile and small IoT devices. The presented framework can be applied to any pixel-wise labeling task such as semantic segmentation. This will be the main topic that we will work on in the future improvements of this work.

\end{document}